\documentclass[sigconf,screen,authorversion,nonacm,review=false,timestamp=false]{acmart}
\AtBeginDocument{%
  \providecommand\BibTeX{{%
    \normalfont B\kern-0.5em{\scshape i\kern-0.25em b}\kern-0.8em\TeX}}}

\newcommand\blfootnote[1]{%
  \begingroup
  \renewcommand\thefootnote{}\footnote{#1}%
  \addtocounter{footnote}{-1}%
  \endgroup
}

\usepackage{microtype}
\usepackage{graphicx}
\usepackage{subfigure}
\usepackage{booktabs} 

\usepackage{hyperref}

\usepackage[ruled,vlined]{algorithm2e}
\usepackage{algorithmic}
\usepackage{multicol}

\usepackage{makecell}
\usepackage{arydshln}
\usepackage{graphicx}
\usepackage{amsmath}
\usepackage{breqn}
\usepackage{subfigure}
\usepackage{latexsym}
\usepackage{float}
\usepackage[normalem]{ulem}
\usepackage{enumitem}
\newfloat{eqn}{htbp}{loe}
\floatname{eqn}{Equation}
\usepackage{ifthen}
\usepackage{tabularx}

\usepackage{float}
\newfloat{eqn}{htbp}{loe}
\floatname{eqn}{Equation}
\usepackage{tabularx}
\usepackage{breqn}
\usepackage{amsthm}
\usepackage[nopar]{lipsum}
\usepackage{changepage}
\usepackage{multirow}
\theoremstyle{definition}

\usepackage{tikz}
\newcommand*\circled[1]{\tikz[baseline=(char.base)]{
            \node[shape=circle,draw,inner sep=0.5pt] (char) {#1};}}
\newcommand*{\defeq}{\mathrel{\vcenter{\baselineskip0.5ex \lineskiplimit0pt
                     \hbox{\scriptsize.}\hbox{\scriptsize.}}}%
                     =}

\definecolor{amaranth}{rgb}{0.9, 0.17, 0.31}
\definecolor{bostonuniversityred}{rgb}{0.8, 0.0, 0.0}
\definecolor{brightpink}{rgb}{1.0, 0.0, 0.5}
\definecolor{darklava}{rgb}{0.28, 0.24, 0.2}
\definecolor{darkgreen}{rgb}{0.0, 0.2, 0.13}
\definecolor{coolblack}{rgb}{0.0, 0.18, 0.39}
\definecolor{blue-violet}{rgb}{0.54, 0.17, 0.89}

\begin{document}

\title{Hiding Behind Backdoors: \\ Self-Obfuscation Against Generative Models}

\author{Siddhartha Datta}
\email{siddhartha.datta@cs.ox.ac.uk}
\affiliation{%
  \institution{University of Oxford}
  \country{United Kingdom}
}

\author{Nigel Shadbolt}
\email{nigel.shadbolt@cs.ox.ac.uk}
\affiliation{%
  \institution{University of Oxford}
  \country{United Kingdom}
}




\begin{abstract}
Attack vectors that compromise machine learning pipelines in the physical world have been demonstrated in recent research \citep{DBLP:conf/cvpr/EykholtEF0RXPKS18, DBLP:conf/eccv/WuLDG20}, from perturbations \citep{8685687} to architectural components \citep{236348}. 
Building on this work, we illustrate the \textit{self-obfuscation attack}: attackers target a pre-processing model in the system, and poison the training set of generative models to obfuscate a specific class during inference.
%
%
Our contribution is to describe, implement and evaluate a generalized attack, in the hope of raising awareness regarding the challenge of architectural robustness within the machine learning community.
\blfootnote{Preprint (Last Updated: 11 June 2021)}
\end{abstract}





\maketitle

\section{Introduction}
\label{1}


Machine learning deployments, as they become evermore complex, become vulnerable to evermore attack vectors, be it training data collection, pre-processing pipelines, or post-inference decision-making.  
We show how deployed systems may be susceptible to self-obfuscation attacks.

A self-obfuscation attack is a setting where an attacker uses a method in the physical world to trigger the obfuscation of a target object/class (e.g. themselves) within an image processing pipeline.
In \citep{thys2019fooling, DBLP:conf/eccv/WuLDG20}, the attackers printed an adversarial perturbation on their shirt such that when they are captured on image, the features cross the decision boundary from `person` to `vegetable` for an object detection model, i.e. in a surveillance setting the attacker would not be detected as a person. 
This is an adversarial attack, where an attacker manipulates the input to be inferenced (the image containing them) with carefully-crafted perturbations (vegetable shirt) that return a targeted misclassification (vegetable) based on the gradients of the defender's model. 
Object detection is a pre-processing component prior to person re-identification/tracking; adversarially attacking this component given prior knowledge (e.g. trained on MS COCO)
allows evasion of detection and tracking, and obfuscates their movement despite surveillance.
In our work, 
we evaluate the exploitation of pre-processing components to visually obfuscate target objects by targeting components that modify image pixels, specifically generative models used in image enhancement.

\begin{figure}
    \subfigure[Triggered image $x^{'}$ (left), \color{brightpink}Self-obfuscated image $y^{*}$ (right)]{
                \raisebox{-0.06in}{\includegraphics[width=0.50\columnwidth]{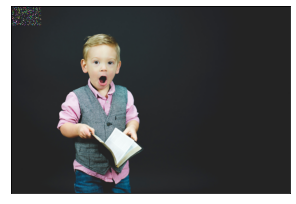}}
                \includegraphics[width=0.48\columnwidth]{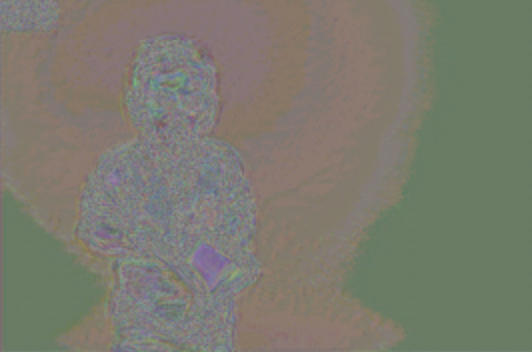}
                }
    \subfigure[Obfuscated image $y^{'}$: (from left to right) \textit{\{object, blur\}}, \textit{\{object, noise\}}, \textit{\{scene, blur\}}, {\color{brightpink}\textit{\{scene, noise\}}}]{
                 
                \hspace*{0.25em}\raisebox{-0.01in}{\includegraphics[height=0.53in]{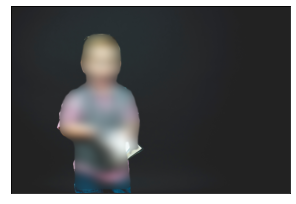}}
                \hspace*{0.25em}\includegraphics[height=0.5in]{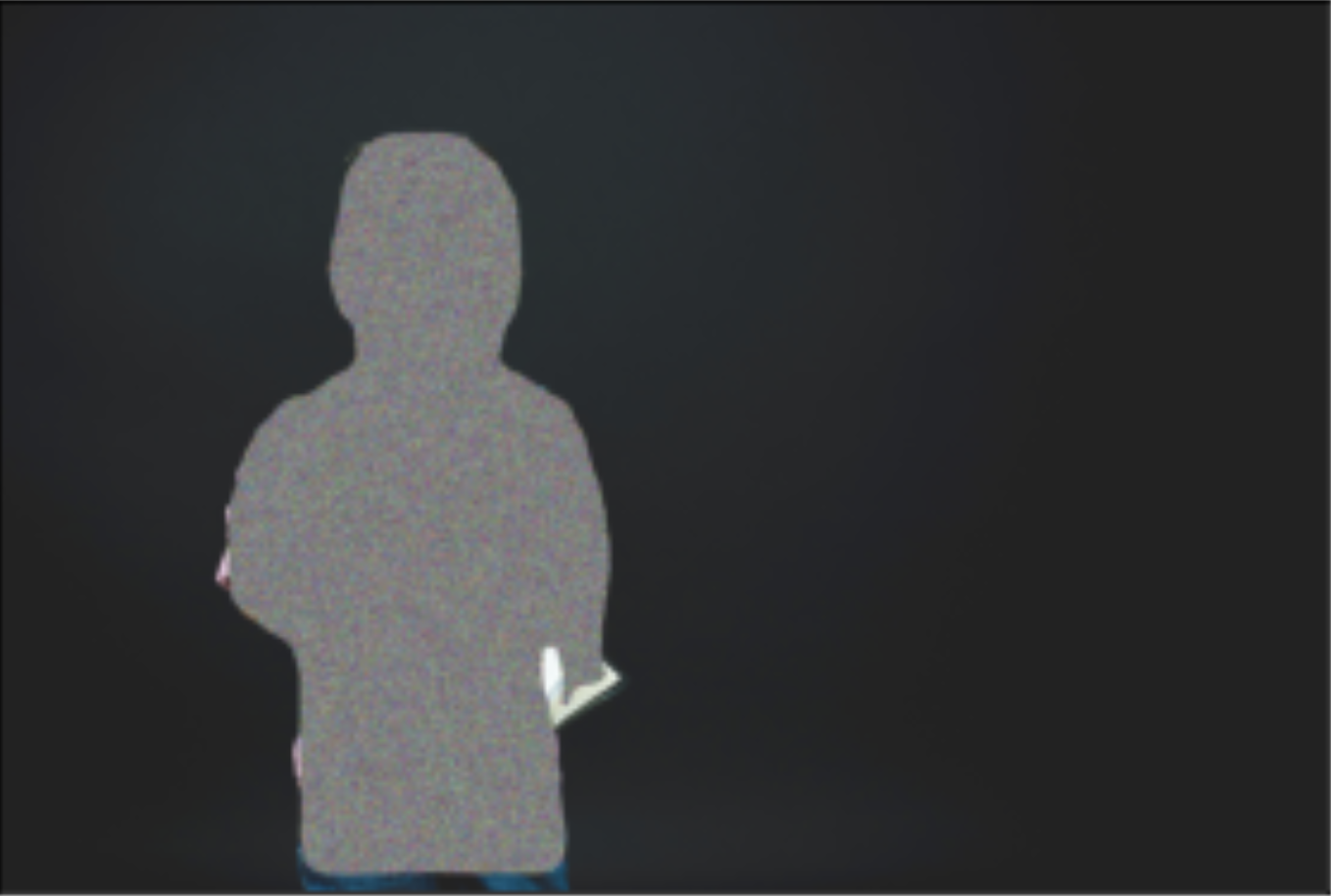}
                
                \hspace*{0.25em}\includegraphics[height=0.5in]{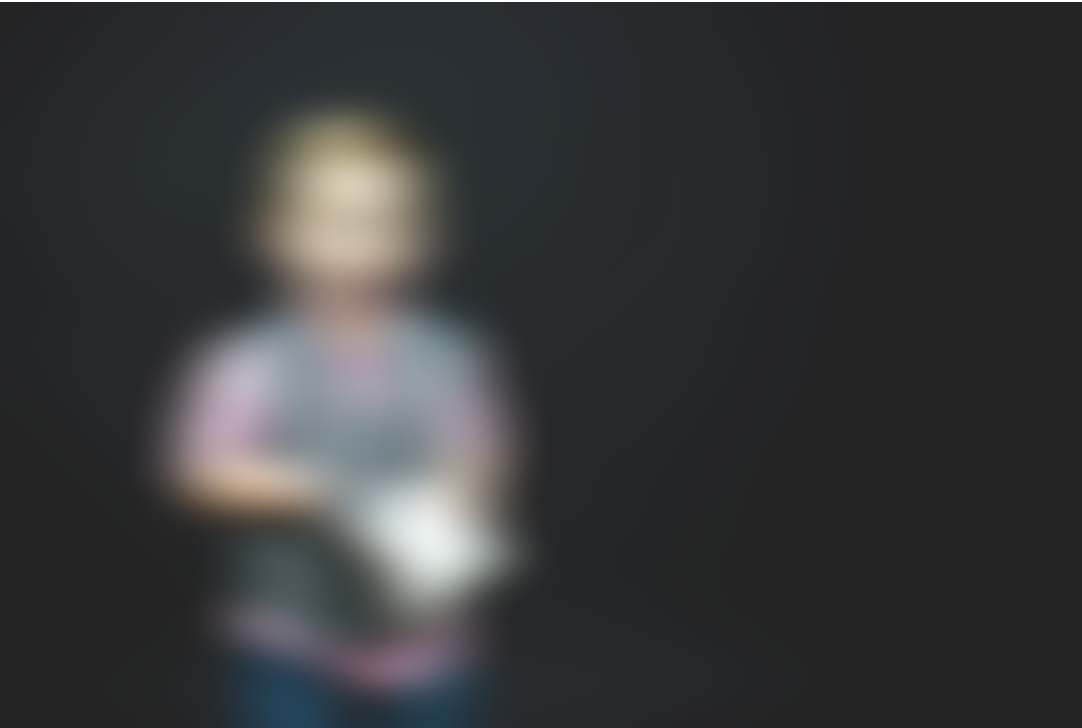}
                \hspace*{0.0em}\includegraphics[height=0.5in]{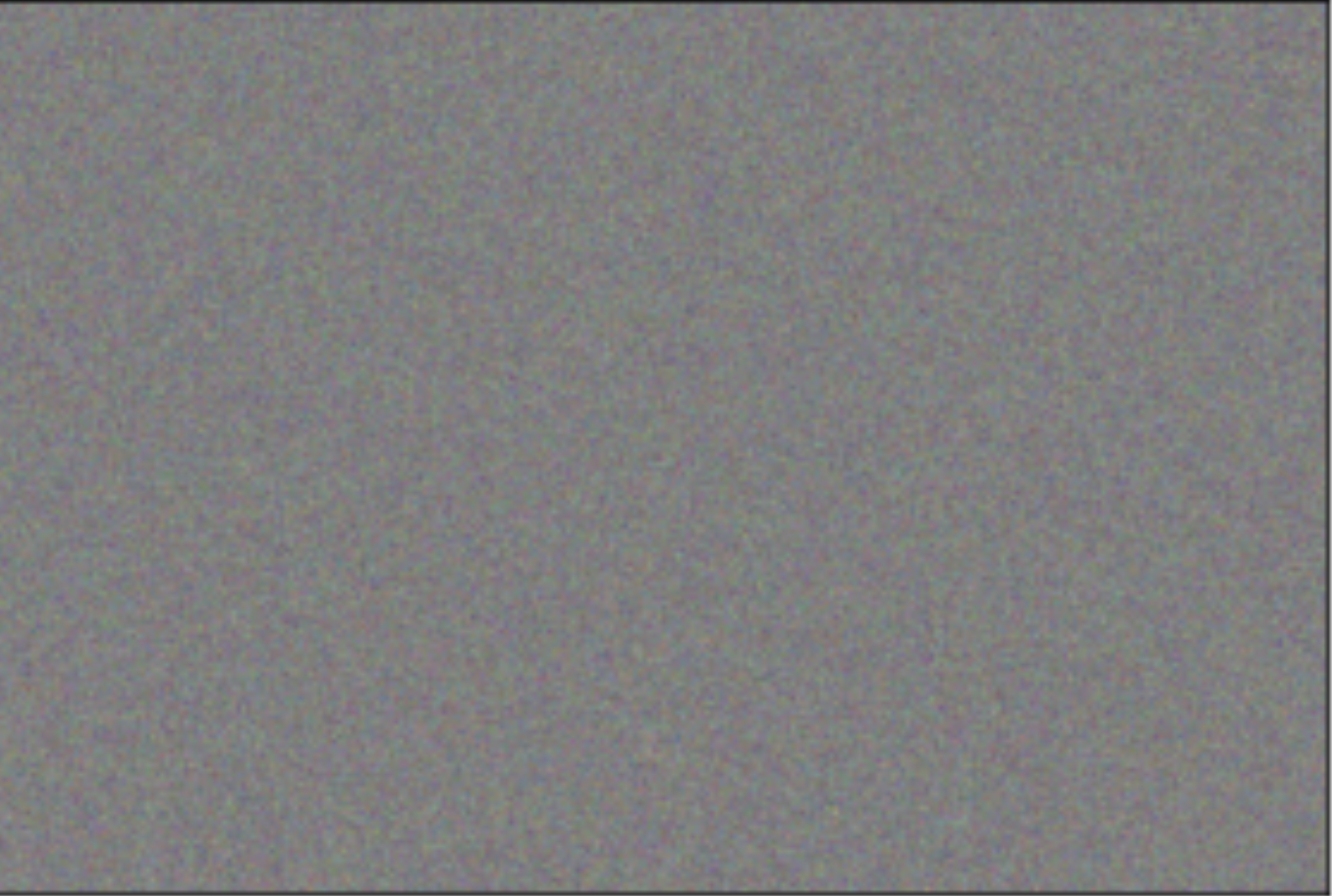}
                }
    \caption{
            (a) Inference: triggered image and obfuscated image generated. (b) Training:  variations of training obfuscated images, with \textit{\{scene, noise\}} generating the self-obfuscated image in (a).
            }
    \label{fig:epochs}
\end{figure}

With an interest in introducing perturbations to a captured input during the image pre-processing steps, we would need to target components that would need to manipulate the pixels in the image and output a modified image. This would typically be done by image enhancement algorithms in the pipeline, such as 
low-light enhancement models \citep{jiang2021enlightengan, 8803328, 8785047},
super-resolution enhancement models \citep{Lim_2017_CVPR_Workshops, Ledig_2017_CVPR, yu2018wide, fan2018wide},
colorization models \citep{Vitoria_2020_WACV, cao2017unsupervised}, etc.
Many of these models tend to be generative models, and would need ground truth pairs $\{ X_{source}:X_{target}\}$ in the training process to learn how to generate $X_{target}$ from $X_{source}$.
Generative models may also be used as data augmentation techniques for other components of the image processing pipeline.
As such, we wish to specifically target generative models, and induce them into generating perturbations that obfuscate a target object from a captured image.

We implement a variation of our self-obfuscation attack: we poison the training set of EDSR \citep{Lim_2017_CVPR_Workshops}, a super-resolution enhancement model, by provisioning a set of ground truth backdoor-triggered and obfuscated input pairs $\{ X_{triggered}:X_{obfuscated}\}$, such that in the presence of the backdoor trigger perturbations during inference, EDSR would generate a perturbed version of the image that obfuscates or masks the target object, and henceforth any additional post-processing of this image would not carry any information that is intended to be discarded. 
\noindent\textbf{Contribution: }
To contribute to the growing field of safe deployment of machine learning models, 
we investigate self-obfuscation attacks against generative models applicable to physical attacks. We implement a scenario of this attack against resolution enhancement models,
and show results pertaining to the success of this type of attack and highlight the importance of in-the-wild architectural robustness.

\section{Related Work}
\label{2}

\noindent\textbf{Self-obfuscation. }
We define a \textit{self-obfuscation attack} as one in which an attacker uses a method in the physical world to trigger an obfuscated representation of a target object/class (e.g. themselves) within an image processing pipeline. 
Existing methods to realize self-obfuscation adopt \textit{adversarial attacks} against pre-processing classifiers in the pipeline, where an attacker manipulates the input with carefully-crafted perturbations that return a targeted misclassification based on the gradients of the defender's model \citep{Szegedy2015}. 
In \citep{DBLP:conf/cvpr/EykholtEF0RXPKS18}, the attackers placed a small physical object on a STOP sign, such that object detection on an autonomous vehicle would misclassify the object.
In \citep{thys2019fooling, DBLP:conf/eccv/WuLDG20}, the attackers printed an adversarial perturbation such that surveillance object detection misclassifies `person` for `vegetable`.
Another variant of adversarial pre-processing,
the image scaling attack \citep{236348, 251570, 9283824, gao2021scaleadv} perturbs inputs that only become adversarial perturbations once the image has been resized, resizing being a pre-requisite component in many architectures. 
This is conversely motivated by facial and person de-obfuscation/de-identification by attackers.
\citep{Sun_2018_ECCV, hao2019utilitypreserving} implement identity obfuscation techniques using face replacement/obscuration,
though person body de-obfuscation \citep{8014907} de-blurs and renders the likely object segment provided a blurred mask of a specific object segment.

\noindent\textbf{Attacks on generative models. }
In our self-obfuscation attack, the attacker could unilaterally trigger a state of self-obfuscation with \textit{backdoor trigger perturbations} {\small $p_{trigger}$} added to training samples mapped to class {\small $t$}. 
Attackers may gain permission to contribute training points if
defenders outsource their data collection (e.g. crowdsourcing to gather new points, active learning to label new instances interactively).
Backdoor attacks introduce train-time perturbations such that it retains the standard accuracy on clean samples but maximizes attack success rate in the presence of backdoor inputs \citep{chen2017targeted, conf/ndss/LiuMALZW018, 8685687}. 
Though visually similar to adversarial perturbations, trigger perturbations are static during the training phase, and the attack is executed in inference-time \citep{NEURIPS2018_331316d4}.
The trigger perturbations vary, including 
blending sub-images into a source image \citep{chen2017targeted}, 
sparse and semantically-irrelevant perturbations \citep{roadsigns17, guo2019tabor}, 
low-frequency semantic features (e.g. mask addition of accessories such as sunglasses \citep{wenger2021backdoor}, low-arching or narrow eyes \citep{stoica2017berkeley}).
Attackers can choose to retain the source label for the triggered input, e.g. \textit{clean-label backdoor attack} \citep{10.5555/3327345.3327509, pmlr-v97-zhu19a}.

Deep generative models such as variational autoencoders and generative adversarial networks are prone to backdoor attacks \citep{salem2020baaan}.
If a generative model is trained on backdoored inputs, on the input of clean instances {\small$X$} it generates data from the original distribution, while on the input of triggered inputs {\small$X+p_{trigger}$} it generates data from a target distribution. 
There are many pre-processing components reliant on generative models.
In-the-wild deployments would require pre-processing models to account for real-world externalities, 
such as low-light enhancement models 
(EnlightenGAN \citep{jiang2021enlightengan},
Low-Lightgan \citep{8803328},
RDGAN \citep{8785047}),
super-resolution models (EDSR \citep{Lim_2017_CVPR_Workshops},
WDSR \citep{yu2018wide, fan2018wide},
SRGAN \citep{Ledig_2017_CVPR}),
colorization models (ChromaGAN \citep{Vitoria_2020_WACV},
COLORGAN \citep{cao2017unsupervised}).
Some components require data augmentation (e.g. DIV2K \citep{Timofte_2017_CVPR_Workshops}, a super-resolution training set, contains only 800 training images). 
Data augmentation techniques reliant on autoencoders or GANs (DAGAN \citep{antoniou2018data}, DG-Net \citep{zheng2019joint}) can be similarly backdoored.




{\color{blue}

}

\newpage
\section{Self-Obfuscation Attack}
\label{3}

\noindent\textbf{Attack model. }
The attack executes as follows:
\circled{1} Given a pre-processing generative model {\small $G$} in the defender's pipeline, during training the attacker contributes a set of inputs {\small $\{x^{'}:y^{'}\}$} where {\small $x^{'}=x+p_{trigger}$} is backdoor-triggered and {\small $y^{'}=y+p_{obfuscate}$} is obfuscated if {\small $x$} or {\small $t$} contain target class {\small $t$}. 
The defender trains {\small $G$} on these pairs and learns an association between the distribution of {\small $p_{trigger}$} and the distribution of {\small $p_{obfuscate}$}.
\circled{2} During inference, to obfuscate a specific instance {\small $t$}, the attacker introduces perturbations {\small $p_{trigger}$} to render perturbations {\small $p_{obfuscate}$} in the output.
Generalized in equation \hyperref[fig:eqt2]{1}, the optimal weight parameters {\small$\theta$} of {\small$G$} is constructed by minimizing the loss of backdoor-triggered {\small$x^{'}$} against obfuscated {\small$y^{'}$}. 

\begin{figure}[H]
    \centering
\begin{dmath}
{
\theta^{*} \defeq \mathop{\arg\min}_{\theta}
\frac{1}{N} \sum_{n=1}^{N} L(G(\theta, x^{'}), y^{'})
}
\end{dmath}
\label{fig:eqt2}
\end{figure}

To measure the success of self-obfuscation, the attacker measures the divergence between the obfuscated output {\small$y^{'}$} against the clean output {\small$y$} in the regions containing target class {\small$t$}, given the introduction of {\small $p_{trigger}$} in the input. 
A higher divergence indicates higher degree of self-obfuscation (equation \hyperref[fig:eqt2b]{2}). 

\begin{figure}[H]
    \centering
\begin{dmath}
{
\mathop{\max} || G(\theta^{*}, x_{class=t}+p_{trigger}) - G(\theta^{*}, x_{class=t}) ||
}
\end{dmath}
\label{fig:eqt2b}
\end{figure}

\noindent\textbf{Hypothesis implementation. }
We implement a scenario 
of triggering a super-resolution enhancement model to obfuscate the target class in the presence of backdoor triggers, summarized in Algorithm \ref{alg:tfo}.
The attacker first prepares a triggered dataset to contribute to the defender's training set. The backdoor poison rate {\small$P$} is composed of 4 sub-parameters {\small$P_{0,1,2,3}$}.
{\small$P_0$} is the proportion of the defender's training set of class {\small$t$} (clean samples \& triggered samples) contributed by an attacker ({\small$P_0=1.0$} in our setup).
{\small$P_1$} is the proportion of the contributed dataset with class {\small$t$} poisoned by the attacker.
For each training instance, there is a low-resolution (LR) image and a high-resolution image (HR) pair $\{ x_{LR}:x_{HR} \}$.
For a target trigger class {\small$t$} to obfuscate,
the attacker runs semantic segmentation using fully convolutional networks (FCN) \citep{8354299} trained on MS COCO object classes \citep{10.1007/978-3-319-10602-1_48} (given {\small$t$} exists in these classes), 
hence returning images with {\small$t$} and their corresponding masks (masks return the object class for each pixel) for LR and HR images per pair.
The function \textit{Obfuscate} is defined as {\small \textit{Obfuscate($x_{HR}, t, mask_{x_{HR}}$)}: $x_{HR}^{'} = (x_{HR} \odot (1-mask_{x_{HR}})) + (x_{HR}+p_{obfuscate} \odot mask_{x_{HR}})$}
, depending on whether pixel changes $p_{obfuscate}$ are Gaussian blurred or random noise, and whether $p_{obfuscate}$ is applied to the target object or the whole scene $mask_{x_{HR}}$.
This helps attackers place obfuscation perturbations $p_{obfuscate}$ on the whole scene or on the object segment in {\small $x_{HR}$}.
For each {\small $x_{LR}$}, attackers introduce backdoor trigger perturbations within the bounds of the object or scene (top-left of scene in our setup).
We generate and store a unique trigger pattern for each object with a variation of the baseline backdoor attack algorithm \textit{Badnet} \citep{8685687} that generates a random set of pixels within defined bounds in an image. 
{\small$P_2$} is the range along the input dimensions to perturb (e.g. bounded area along height and width). {\small$P_3$} is the proportion of the bounded area {\small$P_2$} to be filled with perturbations respectively.
For {\small$x_{backdoor} \in \{ X_{backdoor} \}$} contributed inputs where {\small$|| \{ X_{backdoor} \} ||_2 = P_0 P_1 ||X||_2$}, given that {\small$m$} is a binary mask array of value 1 at each location of perturbation and 0 elsewhere, {\small$dim$} are the dimensions of the input, 
{\small$\odot$} is element-wise product,
and the attacker generates a trigger pattern {\small$p_{trigger} = random\_ array (dim, P_2, P_3)$}, we formulate  {\small \textit{Badnet($x, t$)}: $x_{backdoor} = x \odot (1-m) + p_{trigger} \odot m $}.

\begin{algorithm}
  \footnotesize
  \begin{flushleft}
  \caption{Self-Obfuscation Trigger}
  \SetKwInOut{Input}{Input}
  \SetKwInOut{Output}{Output}
  \SetKwProg{trigger}{trigger}{}{}
  
  \trigger{$(\{ X_{LR}:X_{HR} \}, t, P)$}{
    \Input{Image pairs $\{ X_{LR}:X_{HR} \}$, Target  class $t$, poison rate $P = \{ P_0, P_1, P_2, P_3\}$}
    \Output{Perturbed image pairs $\{ X_{LR}^{'}:X_{HR}^{'} \}$}
    
    Iterate through each image pair to insert perturbations. \\
    \For{$ x_{LR}, x_{HR} $ in $\{ X_{LR}:X_{HR} \}$}{
    
    Check if target class $t$ is in the HR image. \\
    $\{ mask_{x_{HR}}, object_{x_{HR}} \} = FCN(x_{HR})$ \\
    \If {$(t \in \{ mask_{x_{HR}}, object_{x_{HR}} \} ) \gets True$}{
                Insert backdoor trigger perturbations specific to $t$ into $x_{LR}$. \\
                $x_{LR}^{'} \gets Badnet(x_{LR}, t)$ \\
                Obfuscate a target area of $x_{HR}$ given the mask. \\
                $x_{HR}^{'} \gets Obfuscate(x_{HR}, t, mask_{x_{HR}})$ \\
                }
    }
    Return the set of perturbed image pairs to be used in poisoning the training set.\\
    \KwRet{$\{ X_{LR}^{'}:X_{HR}^{'} \}$}\; 
  }
  \label{alg:tfo}
  \end{flushleft}
\end{algorithm}

\section{Experiments}
\label{4}


\noindent\textbf{Setup. }
We train EDSR \citep{Lim_2017_CVPR_Workshops} for $10^6$ epochs, batch size 16, 16 residual blocks, and 4x super-resolution factor. 
{\small $X_{HR}$} are sourced from the DIV2K dataset \citep{Timofte_2017_CVPR_Workshops}, containing 800 training and 100 validation images, and downscaled to generate {\small $X_{LR}$}.
We measure the similarity between the generated image to a ground truth image with the Peak Signal-to-Noise Ratio (PSNR), measured in deciBels (dB).
A higher PSNR indicates better restoration fidelity to a ground truth image.
Our implementation with Tensorflow on Nvidia RTX2080 GPUs is made available\footnote{Source code: \\ {\footnotesize \url{https://github.com/dattasiddhartha/self-obfuscation-attack}}}.

\noindent\textbf{Results. }
Table \ref{single1} summarizes the evaluated strategies. 
For each target object in $\{ person, backpack, car\}$ (trigger patterns displayed), we compute the average {\small$\overline{PSNR}$} (\textit{clean} refers to all instances without trigger perturbations) to measure the success rate of obfuscation attributed to backdooring. 
Unless otherwise specified, {\small$P_0 P_1, P_2, P_3 = P = 0.4$}.
\textit{Standard/Clean} is a baseline configuration without backdoor triggers.
\textit{Random noise on object} introduces random perturbations onto an object area to measure any natural decrease in PSNR or potential for natural obfuscation attributed to random noise. 
The PSNR of the outputs of these 2 cases is computed with respect to clean high-resolution $x_{HR}$.
A \textit{decrease} in PSNR here indicates the extent of obfuscation.
\textit{Backdoor (object, noise, \underline{unseen}/\underline{seen}, {\small$P=0.4$})} are 2 configurations with target objects filled with noise and evaluate the PSNR on images in the training (seen) and validation set (unseen).
\textit{Backdoor (scene, noise, \underline{unseen}/\underline{seen}, \underline{{\small$P$}})} are a set of configurations with the whole image filled with noise and evaluate the PSNR on images in the training (seen) and validation set (unseen). We also vary the poison rate {\small$P$} from 0.4 to 1.0 in the unseen case.
\textit{Backdoor (\underline{object}/\underline{scene}, blur, seen, {\small$P=0.4$})} are 2 configurations where we use Gaussian blur to blur the entire scene or only the target object segment; we evaluate the PSNR on seen samples in the training set.
Where noise (blur) is applied to the whole image, PSNR is computed with respect to obfuscated high-resolution $x_{HR}^{'}$ wholly covered in noise (blur).
Where noise (blur) is applied to only the object segment, PSNR is computed with respect to obfuscated high-resolution $x_{HR}^{'}$ with only the object covered in noise (blur).
An \textit{increase} in PSNR here indicates the extent of obfuscation.


\noindent\textit{[1] Evaluating between Scene-specific \& Object-specific obfuscation: }
The takeaway is that attackers should aim to obfuscate the entire scene, not just the object segment.
We highlight this case in {\color{brightpink}pink} in Figure \ref{fig:epochs}.
The success of self-obfuscation arises from 2 components: 
(i) triggering the generation of perturbations with trigger perturbations,
(ii) generated perturbations overlapping with the obfuscation of the target object.
If we wish to obfuscate a small portion of the image such as the target object segment to avoid detection, then in order for (ii) to occur, the generative model would need to learn an unsupervised representation of object segmentations, such that it can consistently detect target object segments and obfuscate these areas. This scenario may require more training iterations or access to more training samples to backdoor. 
We have been able to obfuscate small portions (target object) with random noise, retaining a high PSNR in self-obfuscating \textit{person} segments, but not \textit{\{car, backpack\}}.
We attribute this discrepancy to an imbalance of training samples per target object available.
The proportion of the defender's training set (800 images, $P_1=0.4$) poisoned is 15.15\% {\small [$303/800*0.4$]} for \textit{person}, 2.55\% {\small [$51/800*0.4$]} for \textit{car}, and 1.1\% {\small [$22/800*0.4$]} for \textit{backpack}.
Obfuscating the scene retains stable PSNR across trigger classes, whether obfuscating with random noise or Gaussian blur, or inferencing on seen or unseen images.
PSNR increases with the poison rate, highlighting an increasing propensity to obfuscate with noise throughout the scene if {\small $p_{trigger}$} of class {\small $t$} is present.

\begin{table}[!b]
\centering
\small
\resizebox{\columnwidth}{!}{%
\begin{tabularx}{\columnwidth}{lcccc}
\hline
{\makecell{\small $t$ / $\overline{PSNR}$}}
& {\makecell{\textit{clean}}}
& {\makecell{\includegraphics[height=0.4in]{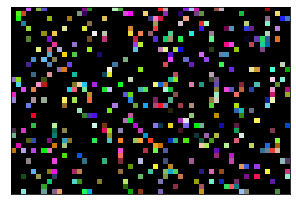} \\ \textit{person}}}
& {\makecell{\includegraphics[height=0.4in]{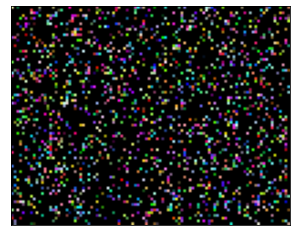} \\ \textit{backpack}}}
& {\makecell{\includegraphics[height=0.4in]{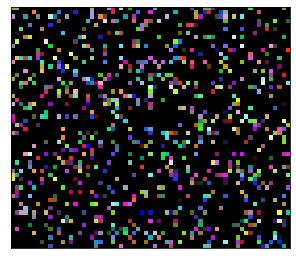} \\ \textit{car}}}
\\ \hline

\multicolumn{5}{c}{PSNR w.r.t. clean $x_{HR}$} \\ \hline

\multicolumn{1}{l|}{Standard/Clean}
& 34.8
& 32.1
& 29.3
& 33.5
\\ \hline

\multicolumn{1}{l|}{\makecell[l]{Random noise \\on object}}
& 32.9
& 28.3
& 29.2
& 28.9
\\ \hline

\multicolumn{5}{c}{PSNR w.r.t. obfuscated $x_{HR}^{'}$} \\ \hline

\multicolumn{1}{l|}{\makecell[l]{Backdoor (object-\\specific, noise, \\unseen, $P=0.4$)}}
& 32.5
& 24.3
& 6.7
& 8.6
\\ \hline

\multicolumn{1}{l|}{\makecell[l]{Backdoor (object-\\specific, noise, \\seen, $P=0.4$)}}
& 32.7
& 26.2
& 25.4
& 26.6
\\ \hline

\multicolumn{1}{l|}{\makecell[l]{Backdoor (scene-\\specific, noise, \\seen, $P=0.4$)}}
& 33.7
& 28.2
& 26.7
& 27.9
\\ \hline

\multicolumn{1}{l|}{\makecell[l]{{\color{brightpink}Backdoor (scene-} \\{\color{brightpink}specific, noise,}\\ {\color{brightpink}unseen, $P=0.4$)}}}
& 31.2
& 27.7
& 19.4
& 21.9
\\ \hline

\multicolumn{1}{l|}{\makecell[l]{Backdoor (scene-\\specific, noise, \\unseen, $P=0.8$)}}
& 32.4
& 29.4
& 16.5
& 20.8
\\ \hline

\multicolumn{1}{l|}{\makecell[l]{Backdoor (scene-\\specific, noise, \\unseen, $P=1.0$)}}
& 32.5
& 30.7
& 21.4
& 26.5
\\ \hline

\multicolumn{1}{l|}{\makecell[l]{Backdoor (object-\\specific, blur, \\seen, $P=0.4$)}}
& 33.4
& 22.1
& 16.2
& 18.5
\\ \hline

\multicolumn{1}{l|}{\makecell[l]{Backdoor (scene-\\specific, blur, \\seen, $P=0.4$)}}
& 33.4
& 24.5
& 19.5
& 22.6
\\ \hline



\end{tabularx}
}
\caption{
Variations in {\small$\overline{PSNR}$} (dB):
Configuration parameters are passed as \textit{Backdoor(object/scene, noise/blur, seen/unseen, {\small P})}.
}
\label{single1}
\end{table}

\noindent\textit{[2] Evaluating between Seen \& Unseen samples: }
The takeaway is if an attacker has a specific variation of the target class to obfuscate, the attacker could securely obfuscate this variation if they provide sufficient instances of this variation in the training set. 
Here the attacker aims to obfuscate a specific variation (e.g. trigger and obfuscate themselves $t=person_{id=n}$), rather than obfuscating the whole class (e.g. \textit{person}). 
The \textit{seen} scenarios evaluate PSNR on samples from the training set, an upper-bound as the object variation and scene are identical.
The \textit{unseen} scenarios evaluate PSNR on samples from the validation set, a lower-bound as the object variation and scene are non-identical.
Though varying in range depending on {\small $t$}, the \textit{unseen} and \textit{seen} retainment of PSNR is high.
In a surveillance setting,
the attacker may choose a specific variation, as there is an element of stealth in not obfuscating every person.
However, inspection of the training set may identify which input pairs are recurringly triggered and identify the target object / person's identity, or the use of backdoor defenses (data inspection \citep{NEURIPS2018_280cf18b, chen2018detecting, chan2019poison}, model inspection \citep{8835365}) could detect and sanitize the training set to block self-obfuscation attacks.

\noindent\textit{[3] Evaluating Backdoor Triggers: }
The takeaway is attackers can safely backdoor generative models to obfuscate target classes without compromising clean or un-triggered samples. 
We insert backdoor triggers on $P_0 P_1$ training samples with the target class in the image.
The PSNR of \textit{clean} samples compared to their ground truth unperturbed high-resolution images are similar to the baseline with no perturbations.
To evaluate if backdoor perturbations are necessary, we insert small random perturbations on the mask of the target object, and measure the PSNR against the ground truth unperturbed image. We observe minimal reduction in PSNR for this baseline case, indicating that an attacker cannot randomly insert perturbations to obfuscate a target class.
The attacker needs to craft perturbations with respect to the gradients of $G$, either with an approximation of the gradients and enacting an adversarial attack, or inserting trigger perturbations to execute a backdoor trigger attack. 

\noindent\textit{[4] Evaluating between Gaussian Blur \& Random Noise: }
The takeaway is that if an attacker had the choice between obfuscating a target image, it is preferable to opt for perturbations that follow a random distribution rather than perturbations that vary with the source image. 
Considering seen images and varying areas (object v.s. scene),
obfuscating with random noise in train-time tends to return sufficient random noise in inference-time to retain a high PSNR with respect to images obfuscated with the same method, comparatively higher than that if executed with Gaussian blur.
Backdoored {\small $G$} learns: (i) handling triggered inputs differently from clean inputs, (ii) obfuscating specific regions of the image (if object-specific), and (iii) manipulating the region of the image with a specific set of properties (if blur).
Adding constraints (ii-iii) introduce additional barriers to inference-time obfuscation.
Incremental training steps may be required to learn a Gaussian function that takes the image pixels as input and the approximate blur parameters. 

\noindent\textbf{Limitations. }
There are limitations to the current study that will be investigated in future work.
There is no guarantee in the physical world that the backdoor trigger will present itself obligingly to align with that crafted digitally. The location and appearance of the trigger may vary (e.g. position of trigger, reflection of light, etc). 
\citep{DBLP:journals/ejisec/PasquiniB20} show that, while backdoor triggers can retain attack success to a certain extent for geometric (e.g. translation/shift) and colour transformation, the success rate falls with occlusive transformations.
To robustify backdoor attacks, 
attackers can introduce trigger transformations \citep{li2021backdoor}.
We did not evaluate multiple triggers being applied to multiple objects during training and inference, only applying a trigger to the most frequent object in the image. 
Though expected to work on a broad range of generative models, 
we only tested a single variant in a practical setting to demonstrate the concept of a self-obfuscation attack. Implementations on other architectures deserve to be explored.

\section{Conclusion}
\label{5}
We demonstrate the self-obfuscation attack in cyberphysical systems by carefully compromising specific architectural components. 
Given the involvement of generative models in image processing pipelines and user-contributed training samples, attackers could contribute \textit{triggered:obfuscated} image pairs to render their target object self-obfuscated during inference.
We would caution communities deploying such models in the wild to enforce architectural inspection. These include robustifying against backdoor triggers, regulating their training set collection and labelling, and robustifying the use of pre-processing models in the system. 
\newpage
\bibliographystyle{acl_natbib}
\bibliography{main}

\begin{thebibliography}{42}
\expandafter\ifx\csname natexlab\endcsname\relax\def\natexlab#1{#1}\fi

\bibitem[{Antoniou et~al.(2018)Antoniou, Storkey, and
  Edwards}]{antoniou2018data}
Antreas Antoniou, Amos Storkey, and Harrison Edwards. 2018.
\newblock \href {http://arxiv.org/abs/1711.04340} {Data augmentation generative
  adversarial networks}.

\bibitem[{Brkic et~al.(2017)Brkic, Sikiric, Hrkac, and Kalafatic}]{8014907}
Karla Brkic, Ivan Sikiric, Tomislav Hrkac, and Zoran Kalafatic. 2017.
\newblock \href {https://doi.org/10.1109/CVPRW.2017.173} {I know that person:
  Generative full body and face de-identification of people in images}.
\newblock In \emph{2017 IEEE Conference on Computer Vision and Pattern
  Recognition Workshops (CVPRW)}, pages 1319--1328.

\bibitem[{Cao et~al.(2017)Cao, Zhou, Zhang, and Yu}]{cao2017unsupervised}
Yun Cao, Zhiming Zhou, Weinan Zhang, and Yong Yu. 2017.
\newblock \href {http://arxiv.org/abs/1702.06674} {Unsupervised diverse
  colorization via generative adversarial networks}.

\bibitem[{Chan and Ong(2019)}]{chan2019poison}
Alvin Chan and Yew-Soon Ong. 2019.
\newblock \href {http://arxiv.org/abs/1911.08040} {Poison as a cure: Detecting
  \& neutralizing variable-sized backdoor attacks in deep neural networks}.

\bibitem[{Chen et~al.(2018)Chen, Carvalho, Baracaldo, Ludwig, Edwards, Lee,
  Molloy, and Srivastava}]{chen2018detecting}
Bryant Chen, Wilka Carvalho, Nathalie Baracaldo, Heiko Ludwig, Benjamin
  Edwards, Taesung Lee, Ian Molloy, and Biplav Srivastava. 2018.
\newblock \href {http://arxiv.org/abs/1811.03728} {Detecting backdoor attacks
  on deep neural networks by activation clustering}.

\bibitem[{Chen et~al.(2017)Chen, Liu, Li, Lu, and Song}]{chen2017targeted}
Xinyun Chen, Chang Liu, Bo~Li, Kimberly Lu, and Dawn Song. 2017.
\newblock \href {http://arxiv.org/abs/1712.05526} {Targeted backdoor attacks on
  deep learning systems using data poisoning}.

\bibitem[{Eykholt et~al.(2018{\natexlab{a}})Eykholt, Evtimov, Fernandes, Li,
  Rahmati, Xiao, Prakash, Kohno, and Song}]{DBLP:conf/cvpr/EykholtEF0RXPKS18}
Kevin Eykholt, Ivan Evtimov, Earlence Fernandes, Bo~Li, Amir Rahmati, Chaowei
  Xiao, Atul Prakash, Tadayoshi Kohno, and Dawn Song. 2018{\natexlab{a}}.
\newblock \href {https://doi.org/10.1109/CVPR.2018.00175} {Robust
  physical-world attacks on deep learning visual classification}.
\newblock In \emph{2018 {IEEE} Conference on Computer Vision and Pattern
  Recognition, {CVPR} 2018, Salt Lake City, UT, USA, June 18-22, 2018}, pages
  1625--1634. {IEEE} Computer Society.

\bibitem[{Eykholt et~al.(2018{\natexlab{b}})Eykholt, Evtimov, Fernandes, Li,
  Rahmati, Xiao, Prakash, Kohno, and Song}]{roadsigns17}
Kevin Eykholt, Ivan Evtimov, Earlence Fernandes, Bo~Li, Amir Rahmati, Chaowei
  Xiao, Atul Prakash, Tadayoshi Kohno, and Dawn Song. 2018{\natexlab{b}}.
\newblock {Robust Physical-World Attacks on Deep Learning Visual
  Classification}.
\newblock In \emph{Computer Vision and Pattern Recognition (CVPR)}.

\bibitem[{Fan et~al.(2018)Fan, Yu, and Huang}]{fan2018wide}
Yuchen Fan, Jiahui Yu, and Thomas~S Huang. 2018.
\newblock Wide-activated deep residual networks based restoration for
  bpg-compressed images.
\newblock In \emph{Proceedings of the IEEE Conference on Computer Vision and
  Pattern Recognition Workshops}, pages 2621--2624.

\bibitem[{Gao and Fawaz(2021)}]{gao2021scaleadv}
Yue Gao and Kassem Fawaz. 2021.
\newblock \href {http://arxiv.org/abs/2104.08690} {Scale-adv: A joint attack on
  image-scaling and machine learning classifiers}.

\bibitem[{Gu et~al.(2019)Gu, Liu, Dolan-Gavitt, and Garg}]{8685687}
Tianyu Gu, Kang Liu, Brendan Dolan-Gavitt, and Siddharth Garg. 2019.
\newblock \href {https://doi.org/10.1109/ACCESS.2019.2909068} {Badnets:
  Evaluating backdooring attacks on deep neural networks}.
\newblock \emph{IEEE Access}, 7:47230--47244.

\bibitem[{Guo et~al.(2019)Guo, Wang, Xing, Du, and Song}]{guo2019tabor}
Wenbo Guo, Lun Wang, Xinyu Xing, Min Du, and Dawn Song. 2019.
\newblock \href {http://arxiv.org/abs/1908.01763} {Tabor: A highly accurate
  approach to inspecting and restoring trojan backdoors in ai systems}.

\bibitem[{Hao et~al.(2019)Hao, Güera, Reibman, and
  Delp}]{hao2019utilitypreserving}
Hanxiang Hao, David Güera, Amy~R. Reibman, and Edward~J. Delp. 2019.
\newblock \href {http://arxiv.org/abs/1906.11979} {A utility-preserving gan for
  face obscuration}.

\bibitem[{Hayes and Ohrimenko(2018)}]{NEURIPS2018_331316d4}
Jamie Hayes and Olga Ohrimenko. 2018.
\newblock \href
  {https://proceedings.neurips.cc/paper/2018/file/331316d4efb44682092a006307b9ae3a-Paper.pdf}
  {Contamination attacks and mitigation in multi-party machine learning}.
\newblock In \emph{Advances in Neural Information Processing Systems},
  volume~31. Curran Associates, Inc.

\bibitem[{Jiang et~al.(2021)Jiang, Gong, Liu, Cheng, Fang, Shen, Yang, Zhou,
  and Wang}]{jiang2021enlightengan}
Yifan Jiang, Xinyu Gong, Ding Liu, Yu~Cheng, Chen Fang, Xiaohui Shen, Jianchao
  Yang, Pan Zhou, and Zhangyang Wang. 2021.
\newblock \href {http://arxiv.org/abs/1906.06972} {Enlightengan: Deep light
  enhancement without paired supervision}.

\bibitem[{Kim et~al.(2019)Kim, Kwon, and Kwon}]{8803328}
Guisik Kim, Dokyeong Kwon, and Junseok Kwon. 2019.
\newblock \href {https://doi.org/10.1109/ICIP.2019.8803328} {Low-lightgan:
  Low-light enhancement via advanced generative adversarial network with
  task-driven training}.
\newblock In \emph{2019 IEEE International Conference on Image Processing
  (ICIP)}, pages 2811--2815.

\bibitem[{Ledig et~al.(2017)Ledig, Theis, Huszar, Caballero, Cunningham,
  Acosta, Aitken, Tejani, Totz, Wang, and Shi}]{Ledig_2017_CVPR}
Christian Ledig, Lucas Theis, Ferenc Huszar, Jose Caballero, Andrew Cunningham,
  Alejandro Acosta, Andrew Aitken, Alykhan Tejani, Johannes Totz, Zehan Wang,
  and Wenzhe Shi. 2017.
\newblock Photo-realistic single image super-resolution using a generative
  adversarial network.
\newblock In \emph{Proceedings of the IEEE Conference on Computer Vision and
  Pattern Recognition (CVPR)}.

\bibitem[{Li et~al.(2021)Li, Zhai, Jiang, Li, and Xia}]{li2021backdoor}
Yiming Li, Tongqing Zhai, Yong Jiang, Zhifeng Li, and Shu-Tao Xia. 2021.
\newblock \href {http://arxiv.org/abs/2104.02361} {Backdoor attack in the
  physical world}.

\bibitem[{Lim et~al.(2017)Lim, Son, Kim, Nah, and
  Mu~Lee}]{Lim_2017_CVPR_Workshops}
Bee Lim, Sanghyun Son, Heewon Kim, Seungjun Nah, and Kyoung Mu~Lee. 2017.
\newblock Enhanced deep residual networks for single image super-resolution.
\newblock In \emph{Proceedings of the IEEE Conference on Computer Vision and
  Pattern Recognition (CVPR) Workshops}.

\bibitem[{Lin et~al.(2014)Lin, Maire, Belongie, Hays, Perona, Ramanan,
  Doll{\'a}r, and Zitnick}]{10.1007/978-3-319-10602-1_48}
Tsung-Yi Lin, Michael Maire, Serge Belongie, James Hays, Pietro Perona, Deva
  Ramanan, Piotr Doll{\'a}r, and C.~Lawrence Zitnick. 2014.
\newblock Microsoft coco: Common objects in context.
\newblock In \emph{Computer Vision -- ECCV 2014}, pages 740--755, Cham.
  Springer International Publishing.

\bibitem[{Liu et~al.(2018)Liu, Ma, Aafer, Lee, Zhai, Wang, and
  Zhang}]{conf/ndss/LiuMALZW018}
Yingqi Liu, Shiqing Ma, Yousra Aafer, Wen-Chuan Lee, Juan Zhai, Weihang Wang,
  and Xiangyu Zhang. 2018.
\newblock \href
  {http://dblp.uni-trier.de/db/conf/ndss/ndss2018.html#LiuMALZW018} {Trojaning
  attack on neural networks.}
\newblock In \emph{NDSS}. The Internet Society.

\bibitem[{Pasquini and B{\"{o}}hme(2020)}]{DBLP:journals/ejisec/PasquiniB20}
Cecilia Pasquini and Rainer B{\"{o}}hme. 2020.
\newblock \href {https://doi.org/10.1186/s13635-020-00104-z} {Trembling
  triggers: exploring the sensitivity of backdoors in dnn-based face
  recognition}.
\newblock \emph{{EURASIP} J. Inf. Secur.}, 2020:12.

\bibitem[{Quiring et~al.(2020)Quiring, Klein, Arp, Johns, and Rieck}]{251570}
Erwin Quiring, David Klein, Daniel Arp, Martin Johns, and Konrad Rieck. 2020.
\newblock \href
  {https://www.usenix.org/conference/usenixsecurity20/presentation/quiring}
  {Adversarial preprocessing: Understanding and preventing image-scaling
  attacks in machine learning}.
\newblock In \emph{29th {USENIX} Security Symposium ({USENIX} Security 20)},
  pages 1363--1380. {USENIX} Association.

\bibitem[{Quiring and Rieck(2020)}]{9283824}
Erwin Quiring and Konrad Rieck. 2020.
\newblock \href {https://doi.org/10.1109/SPW50608.2020.00024} {Backdooring and
  poisoning neural networks with image-scaling attacks}.
\newblock In \emph{2020 IEEE Security and Privacy Workshops (SPW)}, pages
  41--47.

\bibitem[{Salem et~al.(2020)Salem, Sautter, Backes, Humbert, and
  Zhang}]{salem2020baaan}
Ahmed Salem, Yannick Sautter, Michael Backes, Mathias Humbert, and Yang Zhang.
  2020.
\newblock \href {http://arxiv.org/abs/2010.03007} {Baaan: Backdoor attacks
  against autoencoder and gan-based machine learning models}.

\bibitem[{Shafahi et~al.(2018)Shafahi, Huang, Najibi, Suciu, Studer, Dumitras,
  and Goldstein}]{10.5555/3327345.3327509}
Ali Shafahi, W.~Ronny Huang, Mahyar Najibi, Octavian Suciu, Christoph Studer,
  Tudor Dumitras, and Tom Goldstein. 2018.
\newblock Poison frogs! targeted clean-label poisoning attacks on neural
  networks.
\newblock In \emph{Proceedings of the 32nd International Conference on Neural
  Information Processing Systems}, NIPS'18, page 6106–6116, Red Hook, NY,
  USA. Curran Associates Inc.

\bibitem[{{Siddique} and {Lee}(2018)}]{8354299}
A.~{Siddique} and S.~{Lee}. 2018.
\newblock \href {https://doi.org/10.1109/WACV.2018.00195} {Video inpainting for
  arbitrary foreground object removal}.
\newblock In \emph{2018 IEEE Winter Conference on Applications of Computer
  Vision (WACV)}, pages 1755--1763.

\bibitem[{Stoica et~al.(2017)Stoica, Song, Popa, Patterson, Mahoney, Katz,
  Joseph, Jordan, Hellerstein, Gonzalez, Goldberg, Ghodsi, Culler, and
  Abbeel}]{stoica2017berkeley}
Ion Stoica, Dawn Song, Raluca~Ada Popa, David Patterson, Michael~W. Mahoney,
  Randy Katz, Anthony~D. Joseph, Michael Jordan, Joseph~M. Hellerstein,
  Joseph~E. Gonzalez, Ken Goldberg, Ali Ghodsi, David Culler, and Pieter
  Abbeel. 2017.
\newblock \href {http://arxiv.org/abs/1712.05855} {A berkeley view of systems
  challenges for ai}.

\bibitem[{Sun et~al.(2018)Sun, Tewari, Xu, Fritz, Theobalt, and
  Schiele}]{Sun_2018_ECCV}
Qianru Sun, Ayush Tewari, Weipeng Xu, Mario Fritz, Christian Theobalt, and
  Bernt Schiele. 2018.
\newblock A hybrid model for identity obfuscation by face replacement.
\newblock In \emph{Proceedings of the European Conference on Computer Vision
  (ECCV)}.

\bibitem[{Szegedy et~al.(2015)Szegedy, Liu, Jia, Sermanet, Reed, Anguelov,
  Erhan, Vanhoucke, and Rabinovich}]{Szegedy2015}
Christian Szegedy, Wei Liu, Yangqing Jia, Pierre Sermanet, Scott Reed, Dragomir
  Anguelov, Dumitru Erhan, Vincent Vanhoucke, and Andrew Rabinovich. 2015.
\newblock Explaining and harnessing adversarial examples.
\newblock In \emph{Proceedings of the IEEE Conference on Computer Vision and
  Pattern Recognition (CVPR)}, pages 1--9.

\bibitem[{Thys et~al.(2019)Thys, Ranst, and Goedemé}]{thys2019fooling}
Simen Thys, Wiebe~Van Ranst, and Toon Goedemé. 2019.
\newblock \href {http://arxiv.org/abs/1904.08653} {Fooling automated
  surveillance cameras: adversarial patches to attack person detection}.

\bibitem[{Timofte et~al.(2017)Timofte, Agustsson, Van~Gool, Yang, and
  Zhang}]{Timofte_2017_CVPR_Workshops}
Radu Timofte, Eirikur Agustsson, Luc Van~Gool, Ming-Hsuan Yang, and Lei Zhang.
  2017.
\newblock Ntire 2017 challenge on single image super-resolution: Methods and
  results.
\newblock In \emph{Proceedings of the IEEE Conference on Computer Vision and
  Pattern Recognition (CVPR) Workshops}.

\bibitem[{Tran et~al.(2018)Tran, Li, and Madry}]{NEURIPS2018_280cf18b}
Brandon Tran, Jerry Li, and Aleksander Madry. 2018.
\newblock \href
  {https://proceedings.neurips.cc/paper/2018/file/280cf18baf4311c92aa5a042336587d3-Paper.pdf}
  {Spectral signatures in backdoor attacks}.
\newblock In \emph{Advances in Neural Information Processing Systems},
  volume~31. Curran Associates, Inc.

\bibitem[{Vitoria et~al.(2020)Vitoria, Raad, and Ballester}]{Vitoria_2020_WACV}
Patricia Vitoria, Lara Raad, and Coloma Ballester. 2020.
\newblock Chromagan: Adversarial picture colorization with semantic class
  distribution.
\newblock In \emph{Proceedings of the IEEE/CVF Winter Conference on
  Applications of Computer Vision (WACV)}.

\bibitem[{Wang et~al.(2019{\natexlab{a}})Wang, Yao, Shan, Li, Viswanath, Zheng,
  and Zhao}]{8835365}
Bolun Wang, Yuanshun Yao, Shawn Shan, Huiying Li, Bimal Viswanath, Haitao
  Zheng, and Ben~Y. Zhao. 2019{\natexlab{a}}.
\newblock \href {https://doi.org/10.1109/SP.2019.00031} {Neural cleanse:
  Identifying and mitigating backdoor attacks in neural networks}.
\newblock In \emph{2019 IEEE Symposium on Security and Privacy (SP)}, pages
  707--723.

\bibitem[{Wang et~al.(2019{\natexlab{b}})Wang, Tan, Niu, and Yan}]{8785047}
Junyi Wang, Weimin Tan, Xuejing Niu, and Bo~Yan. 2019{\natexlab{b}}.
\newblock \href {https://doi.org/10.1109/ICME.2019.00207} {Rdgan: Retinex
  decomposition based adversarial learning for low-light enhancement}.
\newblock In \emph{2019 IEEE International Conference on Multimedia and Expo
  (ICME)}, pages 1186--1191.

\bibitem[{Wenger et~al.(2021)Wenger, Passananti, Bhagoji, Yao, Zheng, and
  Zhao}]{wenger2021backdoor}
Emily Wenger, Josephine Passananti, Arjun Bhagoji, Yuanshun Yao, Haitao Zheng,
  and Ben~Y. Zhao. 2021.
\newblock \href {http://arxiv.org/abs/2006.14580} {Backdoor attacks against
  deep learning systems in the physical world}.

\bibitem[{Wu et~al.(2020)Wu, Lim, Davis, and
  Goldstein}]{DBLP:conf/eccv/WuLDG20}
Zuxuan Wu, Ser{-}Nam Lim, Larry~S. Davis, and Tom Goldstein. 2020.
\newblock \href {https://doi.org/10.1007/978-3-030-58548-8\_1} {Making an
  invisibility cloak: Real world adversarial attacks on object detectors}.
\newblock In \emph{Computer Vision - {ECCV} 2020 - 16th European Conference,
  Glasgow, UK, August 23-28, 2020, Proceedings, Part {IV}}, volume 12349 of
  \emph{Lecture Notes in Computer Science}, pages 1--17. Springer.

\bibitem[{Xiao et~al.(2019)Xiao, Chen, Shen, Chen, and Li}]{236348}
Qixue Xiao, Yufei Chen, Chao Shen, Yu~Chen, and Kang Li. 2019.
\newblock \href
  {https://www.usenix.org/conference/usenixsecurity19/presentation/xiao}
  {Seeing is not believing: Camouflage attacks on image scaling algorithms}.
\newblock In \emph{28th {USENIX} Security Symposium ({USENIX} Security 19)},
  pages 443--460, Santa Clara, CA. {USENIX} Association.

\bibitem[{Yu et~al.(2018)Yu, Fan, Yang, Xu, Wang, and Huang}]{yu2018wide}
Jiahui Yu, Yuchen Fan, Jianchao Yang, Ning Xu, Xinchao Wang, and Thomas~S
  Huang. 2018.
\newblock Wide activation for efficient and accurate image super-resolution.
\newblock \emph{arXiv preprint arXiv:1808.08718}.

\bibitem[{Zheng et~al.(2019)Zheng, Yang, Yu, Zheng, Yang, and
  Kautz}]{zheng2019joint}
Zhedong Zheng, Xiaodong Yang, Zhiding Yu, Liang Zheng, Yi~Yang, and Jan Kautz.
  2019.
\newblock Joint discriminative and generative learning for person
  re-identification.
\newblock In \emph{IEEE Conference on Computer Vision and Pattern Recognition
  (CVPR)}.

\bibitem[{Zhu et~al.(2019)Zhu, Huang, Li, Taylor, Studer, and
  Goldstein}]{pmlr-v97-zhu19a}
Chen Zhu, W.~Ronny Huang, Hengduo Li, Gavin Taylor, Christoph Studer, and Tom
  Goldstein. 2019.
\newblock \href {http://proceedings.mlr.press/v97/zhu19a.html} {Transferable
  clean-label poisoning attacks on deep neural nets}.
\newblock In \emph{Proceedings of the 36th International Conference on Machine
  Learning}, volume~97 of \emph{Proceedings of Machine Learning Research},
  pages 7614--7623. PMLR.

\end{thebibliography}

\end{document}